\documentclass[10pt,twocolumn,letterpaper]{article}

\usepackage{cvpr}              
\usepackage{multirow}
\usepackage{placeins}
\usepackage{graphicx}
\usepackage{float}
\usepackage{array} 
\usepackage{booktabs} 
\usepackage{makecell} 
\usepackage[table]{xcolor}

\usepackage{framed}
\usepackage{changepage}
\definecolor{formalshade}{rgb}{0.95,0.95,1}
\newenvironment{formal}{%
  \MakeFramed{\advance\hsize-\width\FrameRestore}%
  \noindent\hspace{-4.55pt}
  \begin{adjustwidth}{}{7pt}%
  \vspace{2pt}\vspace{2pt}%
}
{%
  \vspace{2pt}\end{adjustwidth}\endMakeFramed%
}

\usepackage{xspace}

\definecolor{cvprblue}{rgb}{0.21,0.49,0.74}
\usepackage[pagebackref,breaklinks,colorlinks,allcolors=cvprblue]{hyperref}

\def\paperID{18019} 
\def\confName{CVPR}
\def\confYear{2026}

\title{fMRI-LM: Towards a Universal Foundation Model for Language-Aligned fMRI Understanding}

\author{
Yuxiang Wei$^{1,\dagger}$  \quad
Yanteng Zhang$^{1}$ \quad
Xi Xiao$^{2}$ \quad
Chengxuan Qian$^{3}$ \\
Tianyang Wang$^{2}$ \quad
Vince D. Calhoun$^{1,\dagger}$ 
\\[0.3em]
$^{1}$TReNDS Center (Georgia Institute of Technology, Georgia State University, Emory), USA\\
$^{2}$University of Alabama at Birmingham, USA\quad
$^{3}$Jiangsu University, P.R.CHINA\\
{\tt\small $^{\dagger}$weiyuxiang@gatech.edu \quad vcalhoun@gatech.edu}
}

\begin{document}
\maketitle
\begin{abstract}
Recent advances in multimodal large language models (LLMs) have enabled unified reasoning across images, audio, and video, but extending such capability to brain imaging remains largely unexplored. Bridging this gap is essential to link neural activity with semantic cognition and to develop cross-modal brain representations. To this end, we present \textbf{fMRI-LM}, a foundational model that bridges functional MRI (fMRI) and language through a three-stage framework. In Stage~1, we learn a neural tokenizer that maps fMRI into discrete tokens embedded in a language-consistent space. In Stage~2, a pretrained LLM is adapted to jointly model fMRI tokens and text, treating brain activity as a sequence that can be temporally predicted and linguistically described. To overcome the lack of natural fMRI–text pairs, we construct a large descriptive corpus that translates diverse imaging-based features into structured textual descriptors, capturing the \textbf{low-level organization} of fMRI signals. In Stage~3, we perform multi-task, multi-paradigm instruction tuning to endow fMRI-LM with \textbf{high-level semantic} understanding, supporting diverse downstream applications. Across various benchmarks, fMRI-LM achieves strong zero-shot and few-shot performance, and adapts efficiently with parameter-efficient tuning (LoRA), establishing a scalable pathway toward a language-aligned, universal model for structural and semantic understanding of fMRI. Codes and checkpoints are available: https://github.com/yuxiangwei0808/fMRI-LM.
\end{abstract}
\section{Introduction}
\label{sec:intro}

Functional magnetic resonance imaging (fMRI) provides a noninvasive window into human brain activity by capturing blood-oxygen-level-dependent (BOLD) fluctuations across distributed regions. Deep learning has achieved strong performance on supervised fMRI tasks such as phenotype prediction and disease diagnosis~\cite{wei20254d,kim2023swift,kawahara2017brainnetcnn,wangdoctor,li2021braingnn,bedel2023bolt,kan2022brain}, but these models typically require task-specific tuning and labeled data, limiting scalability and cross-study generalization. Recent fMRI foundation models, such as BrainLM and Brain-JEPA~\cite{carobrainlm,dong2024brain}, pretrain on large neuroimaging corpora and transfer well to downstream tasks, yet they remain confined to neural-only objectives (e.g., masked prediction, contrastive learning), requires task-specific tuning, and lack grounding in language.

\begin{figure}
    \centering
    \setlength{\abovecaptionskip}{3pt}
    \includegraphics[width=0.7\linewidth]{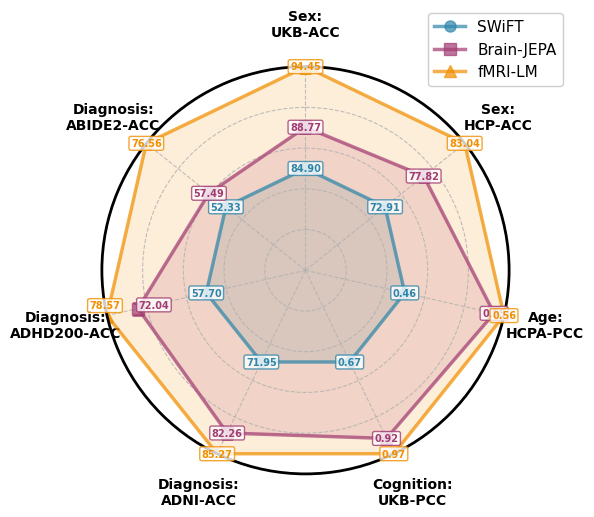}
    \caption{The proposed fMRI-LM outperforms baselines on diverse tasks and shows comprehensive and powerful performance.}
    \label{fig:radar}
\end{figure}

In parallel, large language models (LLMs) and multimodal LLMs (MLLMs) have demonstrated strong cross-modal reasoning over images, audio, and video~\cite{xiao2025visual,brown2020language,li2023blip,li2023llava}. MLLMs typically pair an off-the-shelf LLM with a modality-specific encoder whose outputs are aligned with the text embedding space. Inspired by this design, recent work on EEG \cite{jiang2024neurolm,jiang2024large} treats neural signals as a kind of “language” by quantizing activity into symbolic representations and aligning them with pretrained LLMs. However, these approaches mainly rely on fixed single-question–single-answer templates, underutilizing LLMs’ generative and reasoning capabilities; they also focus on EEG rather than fMRI, and the absence of natural fMRI–text pairs prevents modeling of linguistic semantics that describe brain function. Although several recent works employ LLMs for fMRI-to-text decoding~\cite{qiu2025mindllm,xia2024umbrae,diao2025temporal,diao2026addressing}, they are tailored to task-fMRI settings with explicit stimulus–text pairs and primarily use the LLM’s embedding space to map neural activity back to presented text. In contrast, our goal is to develop a generalizable fMRI foundation model that understands \emph{resting-state and task-independent} neural patterns, without relying on task-evoked paired text.

To explore the potential of LLMs for universal fMRI understanding, we propose \textbf{fMRI-LM}, a foundational model that bridges fMRI and language through a unified multi-stage framework. fMRI-LM is pretrained on over 50{,}000 fMRI scans spanning a wide age range. A key component is a \textbf{structured fMRI–text corpus} that converts imaging-derived features—functional connectivity, graph-theoretical metrics, functional gradients, and ICA (independent component analysis) components—into standardized textual descriptions, providing language-grounded access to the \emph{low-level structure} of fMRI, understood as pre-semantic patterns of connectivity and functional organization analogous to low-level spatial and textural features in images. In Stage~1 (\textbf{fMRI tokenizer training}), a Transformer-based tokenizer with vector quantization maps fMRI into token embeddings aligned with the LLM’s text embedding space. In Stage~2 (\textbf{LLM fine-tuning}), a pretrained LLM is tuned to model fMRI tokens and synthetic fMRI–text pairs, enabling both temporal modeling of brain activity and fMRI-conditioned text generation. Stage~3 (\textbf{downstream instruction tuning}) performs multi-task, multi-paradigm instruction tuning—covering single- and multi-question answering and open-ended description generation—to endow fMRI-LM with \textit{high-level semantic understanding} across diverse neuroscience and clinical tasks. As summarized in Fig.~\ref{fig:radar}, fMRI-LM outperforms strong baselines on a range of benchmarks.

Our key insight is that this descriptive corpus forms a bridge between low-level neural organization and high-level cognitive semantics, analogous to how captions connect image structure to scene meaning in vision–language models. By aligning fMRI with language through this corpus, fMRI-LM learns representations that are transferable across datasets, subjects, and tasks. Overall, our contributions are:
\begin{itemize}
    \item We introduce \textbf{fMRI-LM}, to our knowledge the first LLM-aligned foundational framework for fMRI that maps resting-state and task-independent brain activity into a token space compatible with pretrained language models, enabling a unified interface for fMRI modeling and instruction tuning.
    \item We construct a large-scale \textbf{descriptive corpus} that translates fMRI imaging-based features into structured, caption-like text, providing language supervision that helps the LLM capture the low-level organization and interpretable structure of fMRI signals.
    \item We show that fMRI-LM \textbf{significantly outperforms} supervised and foundation baselines on standard benchmarks, while exhibiting strong \textbf{generalization} across tasks and datasets. Moreover, the model shows notable \textbf{efficiency}, delivering strong results even with limited training data and a small fraction of tunable parameters.
\textbf{}\end{itemize}

\section{Related Work}
\label{sec:related_work}

\noindent\textbf{Brain-LLM Alignment and Convergent Representations:} Recent large-scale analyses reveal that high-performing deep learning models, particularly LLMs, naturally develop representations that align with brain activity. Shen \cite{shen2025alignment} found that brain–model alignment strongly correlates with task performance and even precedes capability gains during training. Likewise, Badr \cite{alkhamissi2025language} suggested that LLMs develop brain-like representations for language and eventually outgrow linguistic rules. Such convergent evolution between biological and language model intelligence indicates that language models may capture representational structures more consistent with human cognition. Motivated by this, we hypothesize that LLMs provide a strong semantic prior for modeling fMRI signals and can enable richer interpretations than task-specific architectures.

\noindent\textbf{Foundation Models for fMRI Understanding:} Recent work in fMRI analysis has shifted from task-specific prediction to general representation learning, driving the development of foundation models that extract transferable neural features from large-scale data. The early supervised CNN and GNN-based approaches performed well in diagnostic tasks but generalized poorly across cohorts \cite{kawahara2017brainnetcnn,li2021braingnn,kan2022brain,kan2022fbnetgen,kim2023swift}. Recent self-supervised models such as BrainLM~\cite{carobrainlm} and Brain-JEPA~\cite{dong2024brain} improve robustness by pretraining with masked reconstruction or contrastive objectives before task-specific fine-tuning. However, these models remain task-bound and lack semantic grounding. Our work addresses this gap by aligning fMRI representations with an LLM backbone to enable unified, language-informed understanding of brain activity.
\section{Methodology}
\label{sec:method}

\begin{figure*}[htbp]
    \centering
    \includegraphics[width=0.99\textwidth]{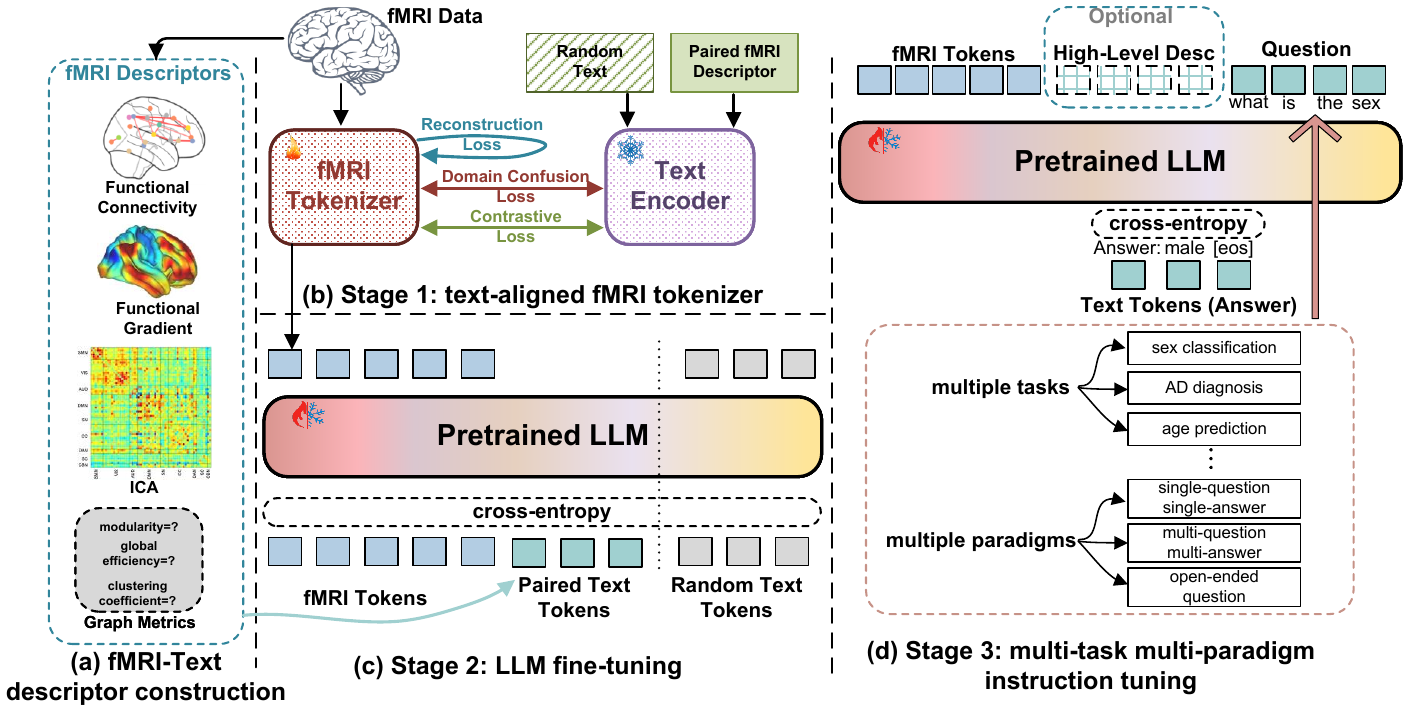}
    \caption{Overall training pipeline of \textbf{fMRI-LM}. 
        (a) fMRI–text pairs are constructed from four types of features: functional connectivity, graph metrics, functional gradients, and ICA-based components. 
        (b) \textit{Stage 1:} align the fMRI tokenizer with the frozen text embedding space. 
        (c) \textit{Stage 2:} tune a pretrained LLM to generate linguistic or temporal representations conditioned on fMRI tokens. Use either full fine-tuning or LoRA \cite{hu2022lora}.
        (d) \textit{Stage 3:} multi-task multi-paradigm instruction tuning for downstream tasks. High-level descriptions are used as optional input for enhanced performance.}
    \label{fig:framework}
\end{figure*}

In this section, we explain the modules and training pipeline of fMRI-LM. As illustrated in \cref{fig:framework}, we first train an fMRI tokenizer composed of a ViT-based encoder \cite{dong2024brain} and a quantizer \cite{van2017neural,mentzer2023finite} that produces fMRI tokens aligned with the frozen text space. A pretrained LLM is then tuned to predict the fMRI tokens and text tokens, followed by supervised instruction tuning.

Given the 4D fMRI $X_{\text{raw}} \in \mathbb{R}^{T \times X \times Y \times Z}$, we follow previous works \cite{dong2024brain,carobrainlm} and parcellate into ROI-level fMRI signals based on atlas Schaefer-400 \cite{schaefer2018local} for cortical regions and Tian-Scale III \cite{tian2020topographic} for subcortical regions, resulting in $N=450$ ROIs: $X \in \mathbb{R}^{T \times N}$.

\begin{table}[t!]
\setlength{\abovecaptionskip}{1pt}
\caption{fMRI-text descriptors}
\label{tab:fmri_desc}
\centering
\resizebox{\linewidth}{!}{
\begin{tabular}{c|c|c}
\hline
type                                 & level  & name                                                    \\ \hline
\multirow{2}{*}{FC}                  & ROI    & Network-pair connectivity                               \\
                                     & Global & Top/bottom connectivity patterns                        \\ \hline
\multirow{3}{*}{FG} & ROI    & Network gradient values                                 \\
                                     & Global & Principal/second/third gradient range                   \\
                                     & Global & Gradient variance                                       \\ \hline
\multirow{4}{*}{ICA}                 & ROI    & Network temporal amplitude, variability, spectral ratio \\
                                     & ROI    & Network-pair FNC                                        \\
                                     & ROI    & Network fALFF                                           \\
                                     & Global & Overall temporal amplitude, variability, spectral ratio \\ \hline
\multirow{2}{*}{Graph}               & ROI    & Network strength                                        \\
                                     & Global & Modularity, global efficiency, average clustering coefficient \\ \hline
\end{tabular}}
\end{table}

\subsection{fMRI-Text Descriptor Construction}
\label{sec:descriptor}
In common vision–language model training, each image is paired with one or more textual descriptions that capture its spatial structure and semantic content. Such captions provide a bridge between visual and linguistic representations, enabling effective multimodal alignment. However, due to the abstract and high-dimensional nature of fMRI data, no analogous text descriptions exist in prior studies. 

\begin{figure}
    \centering
    \setlength{\abovecaptionskip}{3pt}
    \includegraphics[width=0.9\linewidth]{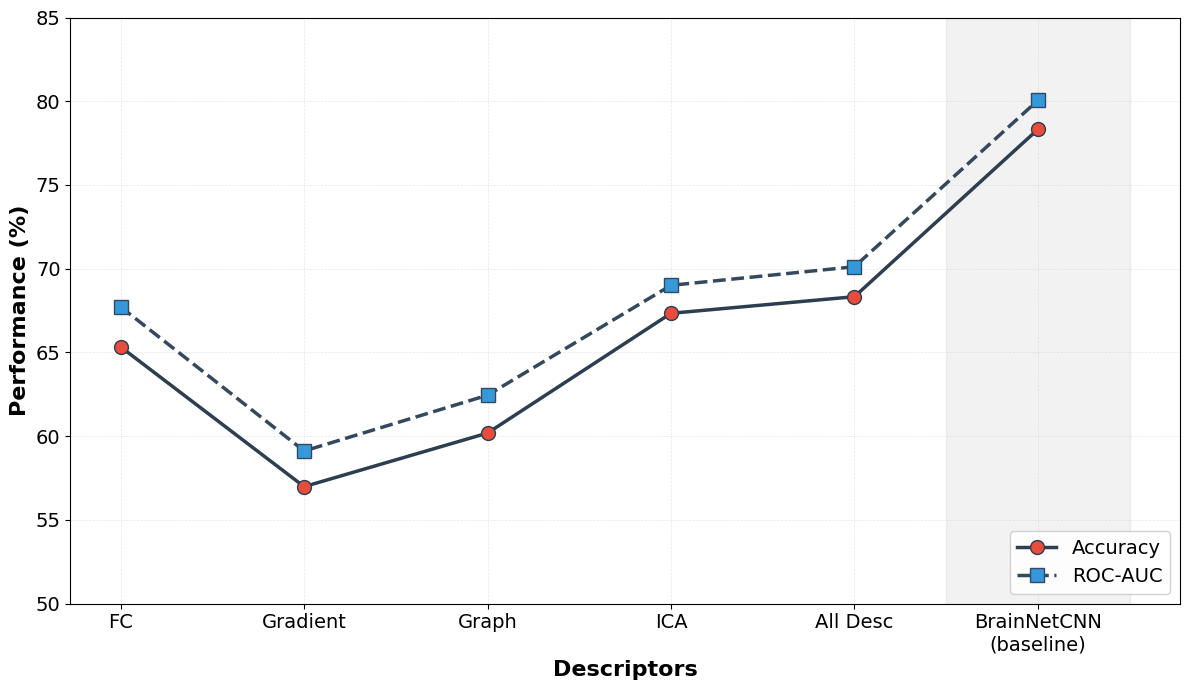}
    \caption{Descriptors' predictive strength over UKB sex. Using all descriptors ("All Desc") can achieve about 70\% accuracy.}
    \label{fig:descrip_perform}
\end{figure}

To address this challenge, we curate a structured text corpus that describes each fMRI data in terms of four complementary feature domains: \textit{functional connectivity (FC)}, \textit{functional gradient (FG)}, \textit{graph-theoretical metrics}, and \textit{independent component analysis (ICA)}. Each description summarizes both the region-of-interest (ROI)–level and global characteristics derived from these representations, as detailed in \cref{tab:fmri_desc}. All functional brain measures are z-scored and normalized relative to the cohort distribution (UK Biobank) to enable interpretable, standardized comparisons across subjects. The quantitative values are then fit to a template to generate cohesive text descriptions. Together, these descriptors capture diverse aspects of intrinsic brain organization and serve as linguistic analogs of neural representations, facilitating multimodal alignment with language models. Complete explanations and the meaning of each descriptor are elucidated in \textbf{Appendix A}. To ensure the descriptors contain meaningful information, we train a BERT classifier \cite{devlin2019bert} with the 4 types of descriptors for UKB sex prediction and compare it with the BrainNetCNN \cite{kawahara2017brainnetcnn} model, as presented in \cref{fig:descrip_perform}.

Subject attributes, such as demographics, are widely used to enhance downstream task accuracy, especially disease diagnosis \cite{wei20254d,xu2025brainprompt}. We further utilize the demographics, phenotypical, cognitive, and physical attributes to construct high-level subject descriptions. These semantic descriptions are only used in Stage 3 disease- and cognition-related tasks. More details are given in \cref{sec:exp_setting} and in \textbf{Appendix A.5}.

\begin{figure}
    \centering
    \setlength{\abovecaptionskip}{3pt}
    \includegraphics[width=0.97\linewidth]{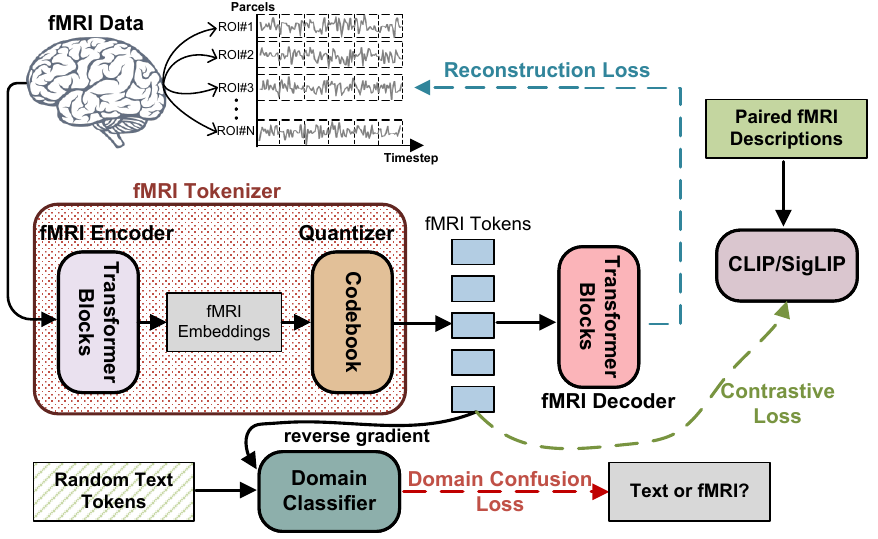}
    \caption{Overview of the fMRI tokenizer, which consists of a Transformer-based encoder and a vector quantizer. The tokenizer is trained with reconstruction, domain-adversarial, and contrastive alignment losses to align fMRI representations with the LLM’s text-embedding space.}
    \label{fig:tokenizer}
\end{figure}

\vspace{-5pt}

\subsection{Text-Aligned fMRI Tokenizer}
To enable pretrained LLMs to understand non-text modalities, it is essential to first encode the input data into embeddings that are \textit{aligned with the frozen text space}. We therefore design a text-aligned fMRI tokenizer, which maps fMRI signals into discrete neural tokens that share a consistent representational geometry with language embeddings.

\vspace{0.3em}
\noindent\textbf{Architecture.}
We employ a Transformer-based encoder $\mathcal{E}_\theta$ inspired by recent neural encoding frameworks~\cite{dong2024brain}, followed by a vector quantizer to discretize the continuous embeddings. As shown in \cref{fig:tokenizer}, given an fMRI input $X \in \mathbb{R}^{T \times N}$, the encoder $\mathcal{E}\theta(\cdot)$ produces a latent feature sequence $z = \mathcal{E}_\theta(X)$, where $z \in \mathbb{R}^{M \times C}$, $C$ denotes the embedding dimension, and $M = [\tfrac{T}{P}] \times N = T' \times N$ is the sequence length. The patch size $P$ is applied only along the temporal dimension to preserve all ROI features.

A quantization module $\mathcal{Q}$, then maps each continuous latent vector $z_m$ to a discrete representation $\widetilde{z}_m$. This module can be implemented using various schemes, such as standard vector quantization (VQ) \cite{van2017neural}. The resulting sequence $\tilde{z} = [\tilde{z}_1, \cdots, \tilde{z}_M]$ serves as discrete fMRI representation.

To preserve information fidelity, a lightweight decoder $\mathcal{D}_\phi$ is trained to reconstruct the original input. The objective for the tokenizer's autoencoding component is:
\begin{equation}
\mathcal{L}{\text{quant}} =
\| X - \mathcal{D}\phi(\tilde{z}) \|_2^2
+ \mathcal{L}_{\text{commitment}},
\label{eq:quant}
\end{equation}
where the first term is the reconstruction loss and $\mathcal{L}_{\text{commitment}}$ is a regularizing term depends on $\mathcal{Q}$.

\noindent\textbf{Domain-Adversarial Alignment.}
Following the domain adaptation strategy of \cite{jiang2024neurolm,ganin2016domain}, we align fMRI embeddings with the text-embedding space of a pretrained LLM. Specifically, we sample text embeddings $z_{\text{text}} \in \mathbb{R}^{M \times C}$ from a frozen LLM (e.g., GPT-2) using tokens drawn from OpenWebText~\cite{Gokaslan2019OpenWeb}. A domain classifier $\mathcal{C}$ is trained to discriminate whether a given embedding originates from fMRI or text, while a gradient reversal layer (GRL) is applied between $\mathcal{E}_\theta$ and $\mathcal{C}$ to enforce confusion. The domain-adversarial loss is defined as
\vspace{-1em}
\begin{equation}
    \mathcal{L}_{\text{domain}} 
    = - \frac{1}{M} \sum_{m=1}^M 
    \Big[ t_m \log \mathcal{C}(z_m) + (1 - t_m) \log (1 - \mathcal{C}(z_m)) \Big]
\label{eq:domain}
\end{equation}
where $t_m = 1$ if the sample is from fMRI and $t_m = 0$ otherwise. This adversarial objective encourages the fMRI tokenizer to produce embeddings that are indistinguishable from text embeddings in the LLM space.

\vspace{0.3em}
\noindent\textbf{Contrastive Cross-Modal Alignment.}
To further bridge the modality gap, we leverage the synthetic text descriptors introduced in \cref{sec:descriptor} to form paired fMRI–text data. We employ a SigLIP-style contrastive loss~\cite{zhai2023sigmoid}, which maximizes similarity between paired embeddings while minimizing it across unpaired samples:
\begin{equation}
    \mathcal{L}_{\text{contrast}} 
    = - \frac{1}{B} \sum_{i=1}^B 
    \log \frac{\exp(\sigma \cdot \text{sim}(z_i, z^{+}_i))}
    {\sum_{j=1}^B \exp(\sigma \cdot \text{sim}(z_i, z^{+}_j))}
\label{eq:contrast}
\end{equation}
where $\text{sim}(\cdot,\cdot)$ denotes cosine similarity, $z_i$ and $z_i^+$ represent matched fMRI and text features, $\sigma$ is a temperature parameter, and $B$ is the batch size.

\vspace{0.3em}
\noindent\textbf{Overall Objective.}
The final loss for the text-aligned fMRI tokenizer combines reconstruction, domain-adversarial, and contrastive terms:
\begin{equation}
    \mathcal{L}_{\text{tokenizer}} = \mathcal{L}_{\text{quant}} + \mathcal{L}_{\text{contrast}} + \lambda \mathcal{L}_{\text{domain}}
\label{eq:total}\end{equation}
where $\lambda$ is empirically set to 0.5. Through this joint optimization, the tokenizer learns discrete neural tokens that both preserve fMRI structure and align closely with the semantic geometry of text embeddings.

\subsection{LLM Fine-Tuning and Temporal Modeling}
\label{sec:llm_tuning}

Given discrete fMRI tokens, we fine-tune a pretrained LLM to model temporal dynamics and generate text. Let $\mathbf{z} = \{ z^{(w,n)} \}$ denote the token sequence, where $w = 1,\dots,T'$ is the time index and $n = 1,\dots,N$ indexes ROIs, and let $I^{(w,n)} \in \mathcal{V}$ be the corresponding vocabulary indices generated from the quantizer.

\vspace{0.3em}
\noindent\textbf{Model Input and Objectives.}
Unlike standard language modeling, where the model predicts the next word in a textual sequence, fMRI data exhibit a \textit{temporal–spatial} structure. Inspired by~\cite{jiang2024neurolm}, we adapt the LLM to perform \textbf{temporal next-step prediction}. Formally, given tokens from $N$ ROIs at time $w$, the LLM predicts the $N$ tokens at $w+1$.

To endow the LLM with both neural and linguistic competence, we introduce three complementary training paradigms as illustrated in \cref{fig:framework}(c):
\begin{itemize}
    \item fMRI-to-fMRI (F2F): temporal next-step prediction of fMRI tokens using \cref{eq:f2f}:
\begin{equation}
\mathcal{L}_{\text{F2F}}
= - \sum_{w=1}^{T'-1}
\sum_{n=1}^{N}
\log P_{\theta}\left( I^{(w+1,n)} \mid \mathbf{z}^{(w,1)}, \dots, \mathbf{z}^{(w,N)} \right)
\label{eq:f2f}
\end{equation}
where $\theta$ denotes the LLM parameters and $P_{\theta}$ represents the autoregressive probability distribution over the extended vocabulary. This objective encourages the LLM to capture temporal dependencies in neural activity.
    \item fMRI-to-Text (F2T): conditioned text generation, where the model learns to produce descriptive text tokens given an fMRI token sequence.
    \item Text-to-Text (T2T): standard language modeling with random text to preserve LLM’s original linguistic ability.
\end{itemize}

The combined loss for LLM alignment is thus:
\begin{equation}
    \mathcal{L}_{\text{LLM}} 
    = \mathcal{L}_{\text{F2T}} 
    + \alpha \mathcal{L}_{\text{F2F}}
    + \beta \mathcal{L}_{\text{T2T}}
\label{eq:llm_total}
\end{equation}
where $\alpha$ and $\beta$  are empirically set to $0.1$ and $0.5$.

\begin{figure}[!t]
    \centering
    \setlength{\abovecaptionskip}{3pt}
    \includegraphics[width=0.75\linewidth]{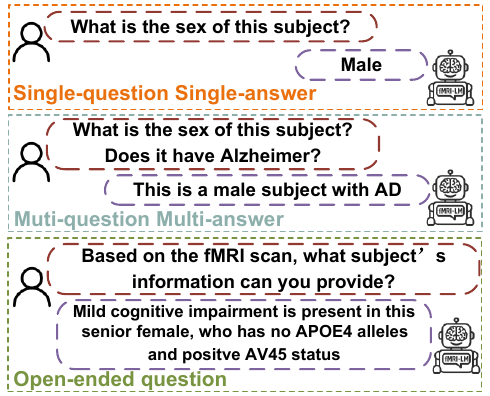}
    \caption{Three paradigms for instruction tuning}
    \label{fig:instruction}
\end{figure}

\subsection{Multi-Task Multi-Paradigm Instruction Tuning}
Foundational vision–language models benefit greatly from diverse instruction-tuning objectives, which enable strong generalization across heterogeneous tasks~\cite{shen2024multitask}. To enable flexible reasoning across diverse neuroscience and clinical tasks, we perform \textbf{multi-task, multi-paradigm instruction tuning} on top of the aligned LLM (Stage~3). Tasks are presented as natural-language queries paired with target responses, covering phenotype prediction, cognitive state classification, and disease diagnosis.

We adopt three paradigms (Fig.~\ref{fig:instruction}): (i) single-question single-answer, (ii) multi-question multi-answer, and (iii) open-ended description. The open-ended setting instructs the model to produce free-form text (e.g., interpreting subject characteristics), encouraging generalizable semantic understanding. Detailed prompt formats and task definitions are provided in \textbf{Appendix D}.

Importantly, the three paradigms are designed primarily as complementary \emph{evaluation} protocols with increasing difficulty, not as a way to enlarge the instruction-tuning corpus. They reuse the same underlying supervision (labels and attributes) but present it under different interaction formats, allowing us to probe how well fMRI-LM handles basic prediction, multi-target reasoning, and free-form generation.

\section{Experiments}
\label{sec:experiment}

\subsection{Experimental Settings}
\label{sec:exp_setting}

\vspace{0.3em}
\noindent\textbf{Datasets.}
We primarily focus on resting-state fMRI due to its wide availability and standardized acquisition protocols. Two large-scale public datasets—UK Biobank (UKB)~\cite{miller2016multimodal} and Adolescent Brain Cognitive Development (ABCD)~\cite{karcher2021abcd}—are used in stage 1 and 2. Each dataset is randomly split with an 80\%--20\% ratio for pretraining and held-out downstream evaluation. To assess generalization and zero-shot transfer, we further include six external datasets spanning multiple age groups and clinical conditions, as summarized in \cref{tab:dataset}.

Since these datasets vary in spatial and temporal resolutions across imaging sites, we standardize all data to ensure consistent temporal resolution and input shape. Specifically, we resample all fMRI time series to a repetition time (TR) of 2.0~s and clip or linearly interpolate each sequence to 160 time points, with 450 ROIs extracted using Schaefer-400 and Tian-Scale III atlases~\cite{schaefer2018local,tian2020topographic}. Before feeding the data into the tokenizer, we perform robust z-score normalization across time for each ROI and apply site-wise variance normalization to mitigate scanner-related biases. 

\begin{table}[t!]
\setlength{\abovecaptionskip}{3pt}
\caption{Dataset summary. Datasets marked with * are also used for zero-/few- shot evaluation. More details in \textbf{Appendix B}.}
\label{tab:dataset}
\centering
\resizebox{\linewidth}{!}{
\begin{tabular}{c|ccc}
\hline
Dataset  & Size  & Age Group & Task     \\ \hline
UKB* \cite{miller2016multimodal}      & 39305 & 37-87     & Sex, Age, Fluid Intel \\
ABCD \cite{karcher2021abcd}    & 18337 & 9-11      & Sex \\
HCP \cite{elam2021human}     & 1079  & 22-37     & Sex, Age, Fluid Comp \\
HCP-A* \cite{elam2021human}   & 632   & 36-100    & Sex, Age, Fluid Comp, Flanker \\
ADNI4 \cite{petersen2010alzheimer}   & 1030  & 55-90     & Sex, Age, AD, Apoe4  \\
ADHD200* \cite{adhd2012adhd} & 624   & 7-22      & ADHD     \\
ABIDE2* \cite{di2017enhancing} & 1114  & 5-64      & Autism   \\ \hline
\end{tabular}%
}
\end{table}


\begin{table*}[htbp]
\renewcommand{\arraystretch}{1.50}  
\setlength{\abovecaptionskip}{3pt}
\caption{Compare fMRI-LM with supervised methods and foundation models on classification tasks. Bold denotes the best method and underline denotes the 2nd best. Note: fMRI-LM-B(G) denotes fMRI-LM-B(GPT2).}
\label{tab:main_result}
\centering
\resizebox{\linewidth}{!}{
\begin{tabular}{c|c|cc|cc|cc|cc|cc}
\hline
\multirow{2}{*}{Type}       & \multirow{2}{*}{Method} & \multicolumn{2}{c|}{UKB-Sex} & \multicolumn{2}{c|}{HCP-Sex} & \multicolumn{2}{c|}{ADNI-AD} & \multicolumn{2}{c|}{ADHD200-ADHD} & \multicolumn{2}{c}{ABIDE2-ASD} \\ \cline{3-12} 
   &        & Acc   & AUC         & Acc     & AUC       & Acc       & AUC       & Acc       & AUC     & Acc       & AUC     \\ \hline
\multirow{5}{*}{Supervised} & BrainNetCNN \cite{kawahara2017brainnetcnn}             & 78.32 (1.12)  & 80.05 (1.06) & 82.01 (2.11)  & 86.94 (1.64) & 75.92 (2.24)  & 77.09 (1.62) & 52.19 (3.14)    & 56.21 (2.82)    & 56.32 (2.45)    & 57.82 (2.51)  \\
                            & BrainGNN \cite{li2021braingnn}                      & 77.31 (2.47)  & 79.53 (1.46) & 79.09 (1.81)  & 81.56 (0.49) & 68.72 (2.84)  & 69.06 (3.11) & 55.67 (2.49)    & 56.72 (1.69)    & 57.02 (2.67)    & 58.09 (3.73)  \\
                            & BNT \cite{kan2022brain}                             & 72.71 (2.64)  & 73.38 (1.92)& 72.67 (1.49)  & 72.05 (1.32)& 70.19 (3.20)  & 72.34 (2.25) & 55.54 (2.39)    & 57.05 (3.32)    & 52.19 (3.10)    & 54.22 (2.44)  \\
                            & FBNETGEN \cite{kan2022fbnetgen}                     & 83.54 (4.67)& 83.56(2.67)& 80.64 (2.41)  & 79.29 (3.91)& 76.74 (2.55)  & 77.64 (1.09)& 49.18 (1.76)    & 52.42 (2.69)    & 51.12 (1.36)    & 54.65 (2.08)  \\
                            & SWiFT \cite{kim2023swift}                           & 84.90 (1.88)& 85.34 (3.19)& 72.91 (1.49)  & 71.69 (1.64) & 71.95 (2.22)  & 70.08 (1.5UY9)& 57.70 (2.44)    & 58.04 (1.88)    & 52.33 (2.21)    & 55.46 (2.72)  \\ \hline
\multirow{3}{*}{Foundation} & BrainLM \cite{carobrainlm}                          & 88.72 (0.88)& 90.42 (0.59) & 81.09 (1.76)  & 82.34 (2.21) & 78.82 (1.54)  & 75.21 (1.68) & 71.22 (1.49)    & 65.21 (1.68)    & {65.22 }(2.28)& {67.29 }(1.18)\\
                            & BrainMass \cite{yang2024brainmass}                  & 92.31 (0.19)  & 92.85 (0.22) & 75.32 (0.49)  & 77.19 (1.01) & 80.05 (2.21)& 83.35 (1.98)& 66.19 (2.27)    & 62.24 (1.79)    & 58.79 (1.12)    & 63.48 (1.79)\\
                            & Brain-JEPA \cite{dong2024brain}                     & 88.77 (0.75)  & 90.13 (0.63) & 77.82 (1.12)  & 79.19 (1.62) & \underline{82.26 }(2.17)& \textbf{84.05 }(2.64)& 72.04 (2.39)    & 65.18 (2.42)    & 57.49 (1.49)    & 64.28 (1.95)  \\ \hline
\rowcolor{cvprblue!15}
                            & fMRI-LM-B(G)    & \textbf{94.89 }(0.22)& \textbf{94.90 }(0.16)& \underline{82.38 }(0.35)& \underline{83.06 }(0.55)& 77.92 (1.01)  & 79.91 (1.25) & \underline{75.06 }(1.06)& \underline{77.14 }(1.97)& \underline{73.44 }(1.21)& \underline{73.02} (1.59)\\
\rowcolor{cvprblue!15}
\multirow{-2}{*}{Ours}      & fMRI-LM-B(Q)    & \underline{94.45 }(0.22)& \underline{94.67 }(0.13)& \textbf{83.04 }(0.26)& \textbf{85.22 }(0.21)& \textbf{85.27 }(0.89)& \underline{81.02 }(1.13)& \textbf{78.57 }(1.24)& \textbf{79.48 }(0.94)& \textbf{76.56} (1.67)& \textbf{76.22} (2.04)\\ 
\hline
\end{tabular}%
}
\end{table*}

\begin{table*}[htbp]
\renewcommand{\arraystretch}{1.50} 
\setlength{\abovecaptionskip}{3pt}
\caption{Compare fMRI-LM with supervised methods and foundation models on regression tasks. Bold denotes the best method and underline denotes the 2nd best. Note: fMRI-LM-B(G) denotes fMRI-LM-B(GPT2).}
\label{tab:main_result_reg}
\centering
\resizebox{\textwidth}{!}{%
\begin{tabular}{c|cccc|cccc|cccc}

\hline
\multirow{3}{*}{Method} & \multicolumn{4}{c|}{UKB}                                                            & \multicolumn{4}{c|}{HCP}                                                          & \multicolumn{4}{c}{HCP-Aging}                                                     \\ \cline{2-13} 
                        & \multicolumn{2}{c|}{Age}                         & \multicolumn{2}{c|}{Fluid Intel} & \multicolumn{2}{c|}{Age}                        & \multicolumn{2}{c|}{Fluid Comp} & \multicolumn{2}{c|}{Flanker}                     & \multicolumn{2}{c}{Fluid Comp} \\ \cline{2-13} 
                        & MAE$\downarrow$& \multicolumn{1}{c|}{p$\uparrow$}            & MAE$\downarrow$             & p$\uparrow$              & MAE$\downarrow$         & \multicolumn{1}{c|}{p$\uparrow$}            & MAE$\downarrow$            & p$\uparrow$              & MAE$\downarrow$          & \multicolumn{1}{c|}{p$\uparrow$}            & MAE$\downarrow$            & p$\uparrow$             \\ \hline
SwiFT                   & 3.40 (0.21)  & \multicolumn{1}{c|}{0.49 (0.073)} & 1.85 (0.092)    & 0.67 (0.011)   & 2.58 (0.25) & \multicolumn{1}{c|}{0.51 (0.046)} & 5.15 (0.39)    & 0.62 (0.11)    & 6.85 (0.44)  & \multicolumn{1}{c|}{0.19 (0.035)} & 5.32 (0.26)    & 0.59 (0.067)  \\ \hline
BrainMass               & 2.01 (0.052) & \multicolumn{1}{c|}{0.77 (0.063)} & 1.59 (0.039)    & 0.88 (0.016)   & 3.01 (0.19) & \multicolumn{1}{c|}{0.49 (0.052)} & 5.29 (0.84)    & 0.55 (0.20)    & 5.66 (0.75)  & \multicolumn{1}{c|}{0.40 (0.016)} & 5.05 (0.49)    & 0.58 (0.16)   \\
Brain-JEPA              & \textbf{1.69 }(0.088)& \multicolumn{1}{c|}{\underline{0.76 }(0.10)}  & 1.59 (0.042)    & \underline{0.92 }(0.011)& \textbf{2.55 }(0.15)& \multicolumn{1}{c|}{\textbf{0.62 }(0.069)} & 4.87 (0.096)   & 0.70 (0.062)   & 5.21 (0.13)  & \multicolumn{1}{c|}{0.44 (0.036)} & 4.88 (0.056)   & \underline{0.73 }(0.074)\\ \hline
\rowcolor{cvprblue!15} fMRI-LM-B(G)    & 1.82 (0.061) & \multicolumn{1}{c|}{\textbf{0.85 }(0.034)} & \underline{1.51 }(0.011)& \textbf{0.95}(0.006)& \underline{2.56 }(0.13)& \multicolumn{1}{c|}{\textbf{0.62 }(0.11)}  & \underline{4.68 }(0.25)& \underline{0.74 }(0.033)& \textbf{5.11 }(0.075)& \multicolumn{1}{c|}{\textbf{0.50 }(0.006)} & \underline{4.70 }(0.29)& \textbf{0.76 }(0.029)\\
\rowcolor{cvprblue!15} fMRI-LM-B(Q)  & \underline{1.80 }(0.056)& \multicolumn{1}{c|}{\textbf{0.85 }(0.029)} & \textbf{1.46 }(0.026)& 0.91 (0.010)& 2.58 (0.14) & \multicolumn{1}{c|}{\underline{0.61 }(0.074)} & \textbf{4.60 }(0.31)& \textbf{0.76 }(0.095)& \underline{5.13 }(0.038)& \multicolumn{1}{c|}{\underline{0.48 }(0.006)} & \textbf{4.61 }(0.26)& \textbf{0.76 }(0.11)\\ 
\hline
\end{tabular}
}
\end{table*}

\noindent\textbf{Paired fMRI-Text Curation and Prompt Expansion.}
To construct paired fMRI–text data, we generate imaging-based textual descriptors for each scan in UKB. Each scan is represented by 23 descriptors derived from four domains, as described in \cref{sec:descriptor}. We use fixed templates to convert numerical statistics into natural-language statements, which are subsequently refined into cohesive paragraphs using DeepSeek-V3~\cite{liu2024deepseek}. Subject-level semantic descriptions are similarly synthesized from demographic and diagnostic attributes to capture broader cognitive or clinical information. More information is in \textbf{Appendix A}.

For instruction tuning, we develop a diverse set of prompts to improve generalization and linguistic robustness. Each downstream paradigm is augmented with up to 200 paraphrased prompt variants generated via LLM rewriting. In training, a random subset of prompts is sampled at each iteration to avoid overfitting to specific phrasings.

\vspace{0.3em}
\noindent\textbf{Implementation Details.}
We introduce three model sizes, \textbf{fMRI-LM-S}, \textbf{fMRI-LM-B}, and \textbf{fMRI-LM-L} with trainable parameters of 46M, 174M, and 610M, respectively (excluding the base LLM). The tokenizer employs a Transformer encoder with a temporal patch size of 32 and a vanilla vector quantizer~\cite{van2017neural}, although other quantizers (e.g., FSQ~\cite{mentzer2023finite}) can be substituted. Unless otherwise stated, we report results for fMRI-LM-B with GPT-2 (124M)~\cite{radford2019language}.

All three stages are trained for 50 epochs using AdamW with a learning rate of $10^{-4}$, cosine-annealing scheduler, and batch size of 32. In Stage~1, we freeze the text encoder and train only the fMRI tokenizer. In Stages~2 and~3, the tokenizer is frozen while the LLM is tuned using either full fine-tuning or parameter-efficient LoRA~\cite{hu2022lora}. More details on training hyperparameters can be found in \textbf{Appendix C}.

\subsection{Main Results}
This part, we systematically evaluate fMRI-LM to answer the following questions: 
\noindent \textbf{Baseline Comparison}: Can fMRI-LM surpass SOTA baselines on standard single-question tasks?
\noindent \textbf{Versatility}: Does instruction tuning allow for diverse formats without performance loss?
\noindent \textbf{Generalization}: Does model demonstrate zero-\noindent \textbf{Efficiency}: Is model effective under data and tuning parameter constraints?

\vspace{0.3em}
\noindent\textbf{Single-Question Single-Answer.}
We first evaluate fMRI-LM-B with GPT-2 and Qwen3-0.6B backbones under the single-question single-answer paradigm, and compare against supervised and foundation models for fMRI. Note that fMRI-LM is tuned jointly on the 5 datasets. For regression targets, the tokenized output space of LLMs is inherently discrete and thus poorly suited for directly predicting continuous values. We therefore adopt two strategies: (i) linear probing~\cite{alain2016understanding} with a lightweight prediction head on top of the LLM’s hidden representations, and (ii) discretizing continuous variables into ordinal bins and formulating them as classification, allowing the model to output faithful discrete responses. Details of the discretization and additional results are provided in \textbf{Appendices B, E, and F}.

As shown in \cref{tab:main_result} and \cref{tab:main_result_reg}, the two variants of fMRI-LM achieve the best or second-best performance on most datasets and targets. Although fMRI-LM underperforms Brain-JEPA on ADNI-AD, we emphasize that the key strength of our framework is its unified instruction-tuning pipeline and its ability to handle diverse tasks without extensive task-specific fine-tuning.

\begin{figure}
    \centering
    \includegraphics[width=0.98\linewidth]{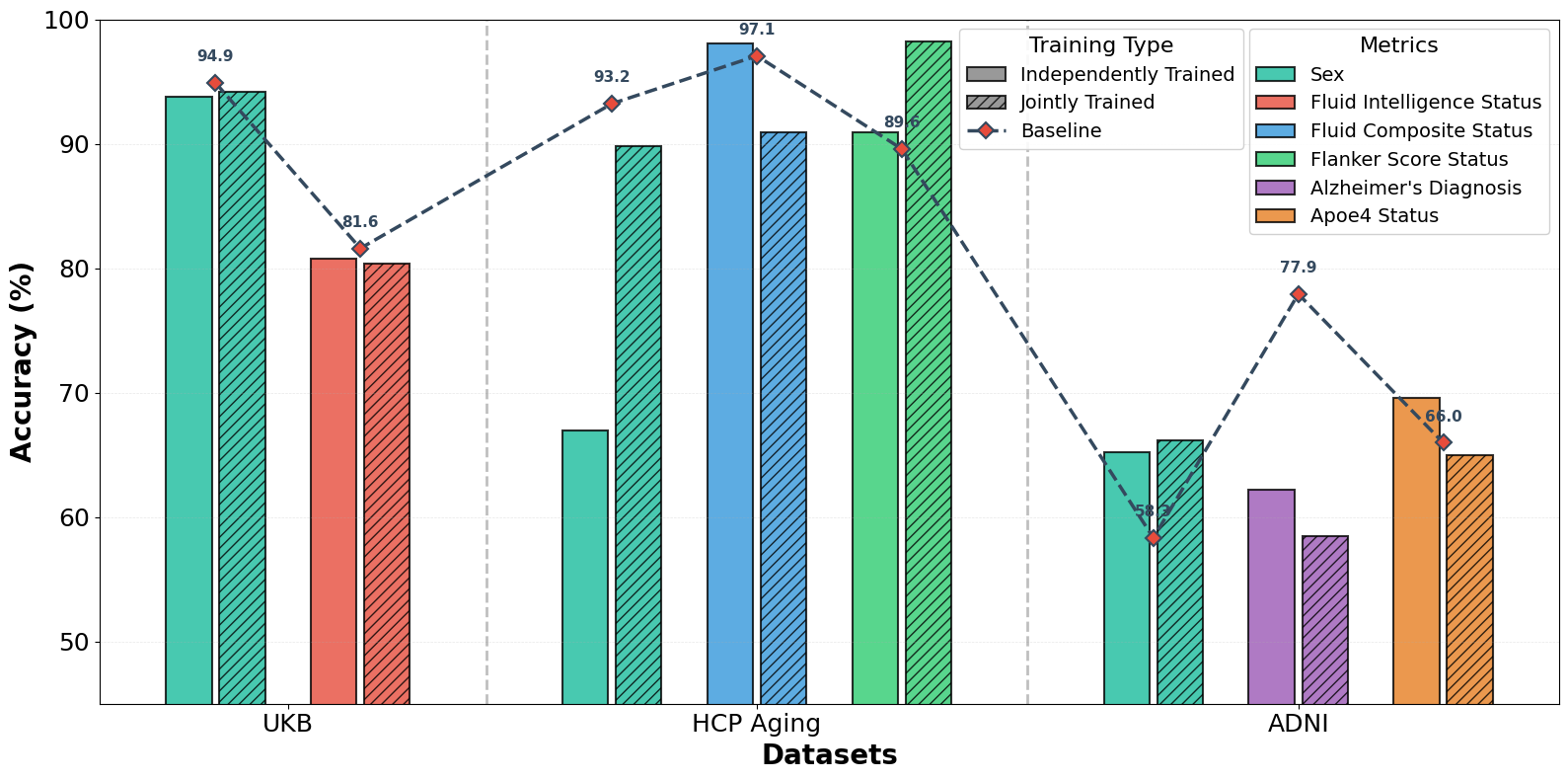}
    \caption{Performance of fMRI-LM on the multi-question multi-answer across UKB, HCP-A, and ADNI. We report results when training independently on each dataset and when jointly training on all three. "baseline" refers to single-question single-answer.}
    \label{fig:metric_mq}
\end{figure}

\begin{figure}
    \centering
    \includegraphics[width=0.95\linewidth]{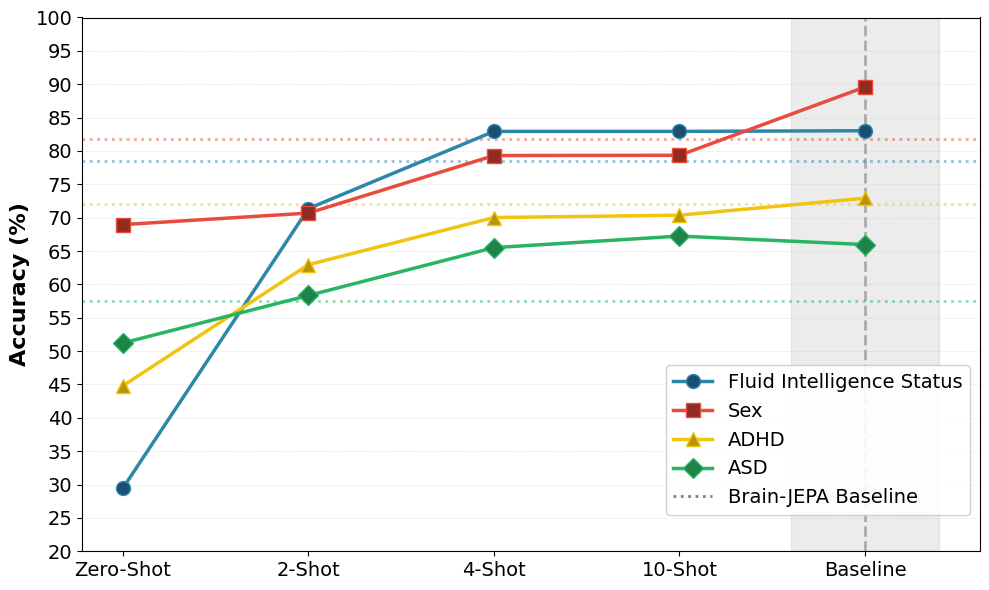}
    \caption{Zero-shot and few-shot generalization of fMRI-LM. We evaluate three settings: new task on the same dataset, same task on a new dataset, and new task on a new dataset. "baseline" refers to single-question single-answer.}
    \label{fig:fewshot}
\end{figure}

\vspace{0.3em}
\noindent\textbf{Multi-Question Multi-Answer.}
We next evaluate the multi-question multi-answer paradigm, where each fMRI scan is paired with multiple questions and the model must predict all answers simultaneously. We report results on UKB, HCP-A, and ADNI, as summarized in \cref{fig:metric_mq}. Detailed definitions of each target and its possible values are provided in \textbf{Appendix D.2}.

Compared to the single-question setting, performance under the multi-question paradigm degrades only marginally relative to strong baselines, with the largest drop observed for AD prediction on ADNI. In contrast, fMRI-LM achieves comparable or even higher accuracy on several targets (e.g., sex, fluid composite, and flanker scores), suggesting that jointly training on multiple, potentially correlated targets can help the model acquire more universal fMRI representations and solve a broader set of tasks.

\begin{figure*}[htbp]
    \centering
    \includegraphics[width=0.87\textwidth]{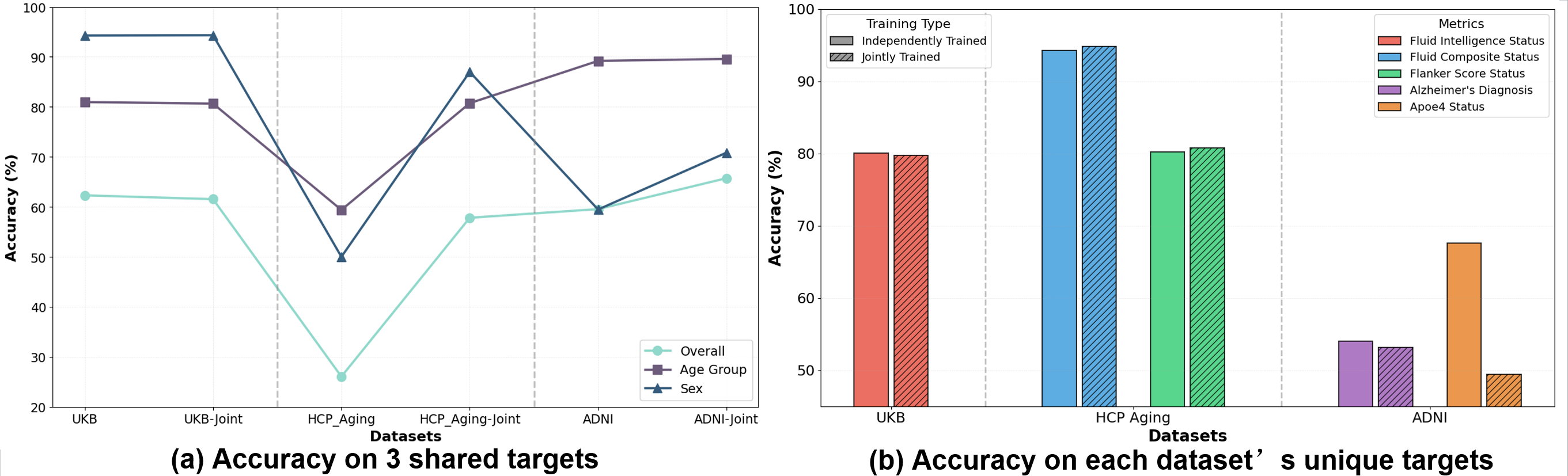}
    \caption{Open-ended question performance of fMRI-LM on UKB, HCP-A, and ADNI. We show models trained independently on each dataset as well as a jointly trained model. ``Baseline'' indicates performance on each target under the single-question single-answer setting.}
    \label{fig:metric_open}
\end{figure*}

\noindent\textbf{Open-Ended Question.}
We further evaluate fMRI-LM on open-ended questions on UKB, HCP-A, and ADNI, as shown in \cref{fig:metric_open}. All datasets share three common target fields, and each dataset additionally includes a small set of unique targets. The “overall” metric requires all fields in a generated sentence to match the ground-truth labels in order for the prediction to be counted as correct. Detailed definitions are given in \textbf{Appendix D.3}.

Since the model is tuned to generate cohesive free-form text instead of structured single or multiple answers, we employ DeepSeek-V3 as an automatic evaluator to determine whether the generated answer matches the target fields, and subsequently perform manual verification by human experts. From \cref{fig:metric_open}, fMRI-LM performs surprisingly well on several targets such as sex and fluid composite status, achieving accuracy comparable to the structured paradigms. Jointly training across datasets (except for ADNI, likely due to its biomarker-driven, disease-specific distribution differing from the others) further improves performance, indicating that fMRI-LM benefits from task and dataset diversity and can develop more universal fMRI understanding.

\vspace{0.3em}
\noindent\textbf{Zero-Shot and Few-Shot Generalization.}
We finally test whether fMRI-LM can generalize to unseen tasks or datasets with no or limited labeled data (2, 4, and 10 samples, balanced by label). We explore three configurations: (i) new task on the same dataset, (ii) same task on a new dataset, and (iii) new task on a new dataset. Concretely, we first train the model on UKB for sex classification, then evaluate its zero-shot and few-shot performance on fluid intelligence status prediction in UKB, sex classification on HCP-A, disease-related tasks on ADHD200 and ABIDE2.

As shown in \cref{fig:fewshot}, fMRI-LM performs relatively poorly in the strict zero-shot setting, but its performance improves substantially even with only two labeled samples. This suggests that the model learns general fMRI representations can be quickly adapted with minimal supervision. Notably, fMRI-LM attains performance comparable to using the full downstream training set under a 4-shot setting for fluid intelligence status prediction and ASD prediction, indicating that the model does not rely heavily on large downstream datasets and can flexibly adapt to a wide variety of tasks.

\subsection{Ablation Studies}
More ablations on model size and loss configurations are provided in \textbf{Appendix~G}.

\begin{figure}[t]
    \centering
    \setlength{\abovecaptionskip}{3pt}
    \begin{minipage}{0.49\linewidth}
        \centering
        \includegraphics[width=\linewidth]{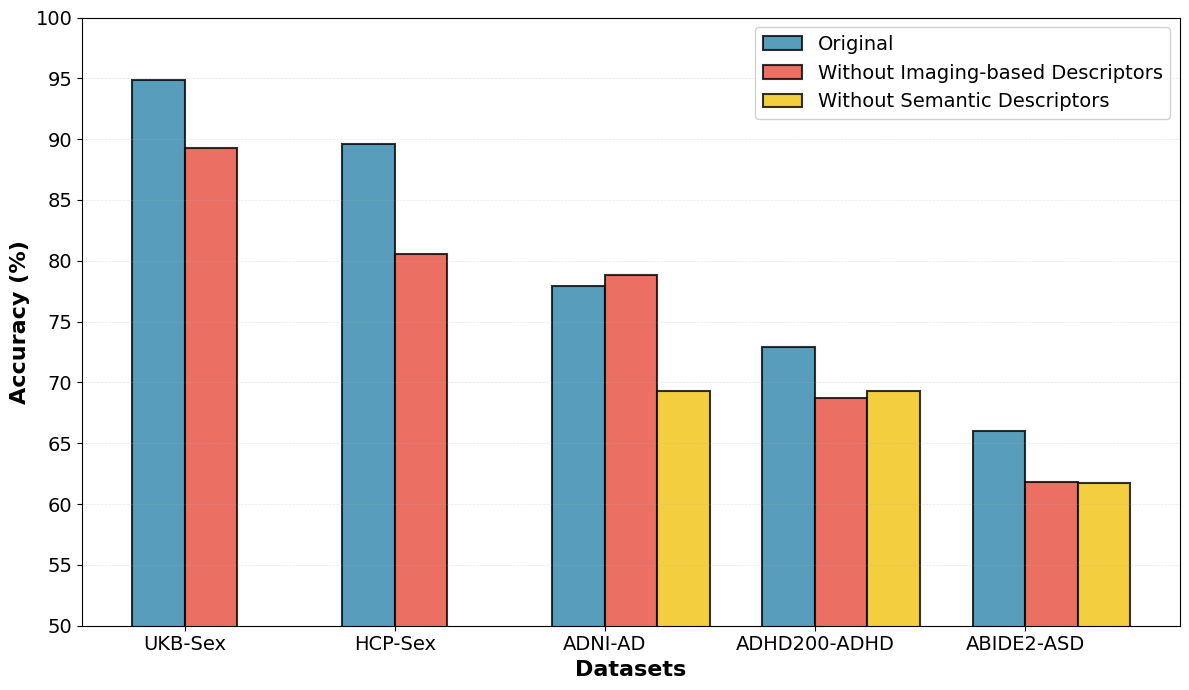}
        \vspace{1pt}
        {\scriptsize (a)}
        \label{fig:ablate_desc}
    \end{minipage}
    \hfill
    \begin{minipage}{0.5\linewidth}
        \centering
        \includegraphics[width=\linewidth]{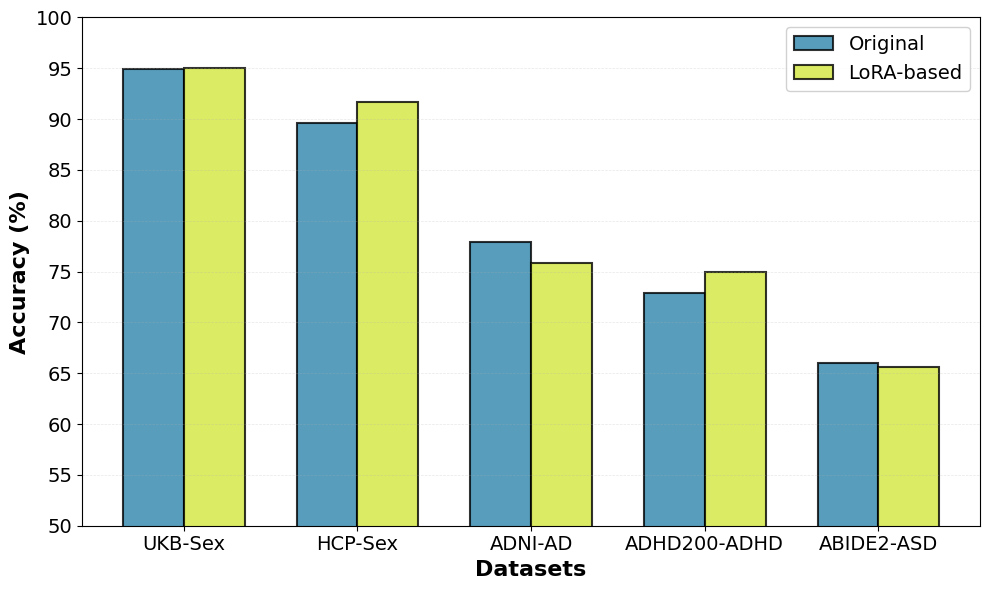}
        \vspace{1pt}
        {\scriptsize (b)}
        \label{fig:ablate_lora}
    \end{minipage}

    \caption{(a) Effect of imaging-based and semantic descriptors. Semantic descriptors are only used for disease- and cognition-related tasks. (b) Effect of LoRA-based tuning on fMRI-LM.}
    \label{fig:ablation_two}
\end{figure}

\vspace{0.3em}
\noindent\textbf{Effect of Imaging-Based and Semantic Descriptors.}
To assess the contribution of the imaging-based descriptors introduced in \cref{sec:descriptor} and \textbf{Appendix~A}, we evaluate fMRI-LM without any paired fMRI–text data during pretraining (i.e., removing the contrastive loss in \cref{eq:contrast} and the F2T objective in \cref{sec:llm_tuning}). We also ablate the high-level semantic text descriptions used as complementary input during downstream tuning. The results are shown in \cref{fig:ablation_two}(a). Removing the imaging-based descriptors—and thus the fMRI–text pairs used during pretraining—significantly degrades performance, especially on sex classification. While performance on ADNI-AD slightly improves after removal, we hypothesize this is due to distributional differences between the pretraining data (UKB/ABCD) and ADNI’s disease-focused population, which may reduce the benefit of descriptor-based alignment in this specific case.

\vspace{0.3em}
\noindent\textbf{Parameter-Efficient Tuning via LoRA.}
While full fine-tuning of LLMs is effective, it can be computationally demanding and prone to overfitting with limited data. To address this, we investigate a parameter-efficient fine-tuning approach using Low-Rank Adaptation (LoRA) \cite{hu2022lora} for Stages 2 and 3. As shown in \cref{fig:ablation_two}(b), employing LoRA not only maintained but in some cases, improved performance on tasks such as HCP sex classification and ADHD diagnosis. This suggests that LoRA effectively adapts the model to fMRI data by tuning only a small fraction of its parameters \cite{xiao2026not}. This approach preserves the rich linguistic knowledge encoded in the pretrained LLM, which is crucial for strong performance, while efficiently learning the relevant neuro-semantic representations from the fMRI inputs.

\begin{figure}
    \centering 
    \setlength{\abovecaptionskip}{3pt}
    \includegraphics[width=1\linewidth]{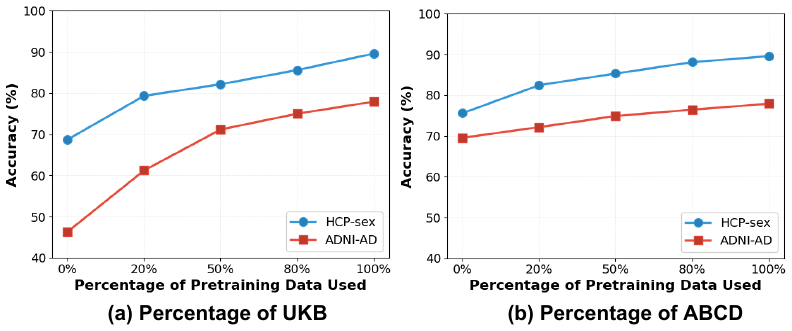}
    \caption{Effect of pretraining data size on downstream performance for HCP sex classification and ADNI AD prediction.}
    \label{fig:scale_datasize}
\end{figure}

\vspace{0.3em}
\noindent\textbf{Impact of Pretraining Data Size.}
To explore the effect of pretraining data scale, we vary the fraction of UKB and ABCD used for pretraining (from 0\% to 100\%) and evaluate downstream performance on HCP sex classification and ADNI AD prediction. As shown in \cref{fig:scale_datasize}, even without UKB or ABCD, the model achieves reasonable performance (around 70\% for sex and 50\% for AD). Removing ABCD has smaller impact than removing UKB, which may be attributed to domain shift (ABCD focuses on children, while most other datasets focus on adults). Overall, performance improves consistently with more pretraining data.

\section{Conclusion}

We introduced fMRI-LM, a foundational framework for universal fMRI understanding that aligns fMRI with LLMs through the synthetic fMRI–text descriptor corpus, which provides scalable linguistic supervision in the absence of natural fMRI–text pairs. Extensive experiments across seven datasets demonstrate fMRI-LM's strong performance and generalization, with further ablation studies validating the importance of paired descriptors and confirming its adaptability and scalability, even under parameter-efficient (LoRA) and few-shot settings. highlighting its adaptability and scalability. This work  presents a step toward unified, language-grounded brain modeling. By leveraging the structure and reasoning capabilities of LLMs, it offers a scalable way to interpret fMRI, integrate heterogeneous tasks, and transfer knowledge across studies.
\FloatBarrier
{
    \small
    \bibliographystyle{ieeenat_fullname}
    \bibliography{main}
}

\clearpage 
\onecolumn 
\appendix

\section{Detailed Explanations on fMRI Descriptors}
This appendix summarizes the full set of imaging-based descriptors used in our analyses. 
For each metric, we provide both ROI-level and global-level definitions along with a brief 
description of the information captured by the feature. Most of the features are based on the Schaefer400 atlas with 7 functional networks: visual, somatomotor, dorsal attention, ventral attention, limbic, fronto-parietal control, and default-mode.

\subsection{Functional Connectivity (FC) Descriptors}
Functional connectivity captures the statistical synchronization between brain regions, reflecting how distributed neural populations co-fluctuate over time. It provides insight into large-scale communication patterns, network integration/segregation, and the overall organizational architecture of intrinsic brain activity.

\subsubsection*{ROI-Level}
\begin{itemize}
    \item \textbf{Network-Pair Connectivity}: Mean correlation strength between predefined 
    network pairs (e.g., Default--Visual, Control--Limbic), z-scored relative to the cohort.  
    \textit{Captures directed pairwise interaction strength among large-scale systems.}
\end{itemize}

\subsubsection*{Global-Level}
\begin{itemize}
    \item \textbf{Top/Bottom Connectivity Patterns}: The three most elevated and three most 
    reduced network-pair connections across the whole brain.  
    \textit{Summarizes extremes of functional integration and dissociation.}
\end{itemize}

\subsubsection*{Example:} The subject's functional connectivity profile reveals several notable patterns:- Among the most pronounced increases: SalVentAttn-SomMot coupling shows mild enhancement (z=+1.34), DorsAttn network exhibits heightened within-network connectivity (z=+1.38), and SalVentAttn-Vis interaction is moderately elevated (z=+1.18).- Regarding diminished connections: Cont-Limbic functional coupling falls within normative bounds (z=-0.85), DorsAttn-Limbic interaction remains in the typical range (z=-0.74), and Default-Limbic connectivity similarly shows no significant deviation (z=-0.64).

\subsection{Functional Gradients Descriptors}
Functional gradients describe the continuous, hierarchical organization of the cortex from low-level sensory processing to high-level transmodal cognition.

\subsubsection*{ROI-Level}
\begin{itemize}
    \item \textbf{Network Gradient Values}: Mean gradient values for each of the seven canonical 
    networks (Control, Default Mode, Dorsal Attention, Ventral Attention, Limbic, Somatomotor, 
    Visual).  
    \textit{Indexes each network's position along macroscale functional gradients.}
\end{itemize}

\subsubsection*{Global-Level}
\begin{itemize}
    \item \textbf{Principal Gradient Range}: Degree of separation between sensory and transmodal 
    regions along the first gradient axis.  
    \textit{Reflects hierarchical organization of perception-to-association cortex.}

    \item \textbf{Second Gradient Range}: Extent of segregation among distinct sensory 
    modalities.  
    \textit{Quantifies differentiation within sensory processing streams.}

    \item \textbf{Third Gradient Range}: Degree of distinction between control and association 
    systems.  
    \textit{Captures higher-order specialization across cognitive control networks.}

    \item \textbf{Gradient Variance}: Overall dispersion/spread of functional gradients.  
    \textit{Represents global heterogeneity of cortical functional organization.}
\end{itemize}

\subsubsection*{Example:} The functional gradient profile for this subject reveals:- Principal gradient extent falls within typical limits (+0.3 SD)- Second gradient span remains within normative bounds (+0.3 SD)- Third gradient coverage aligns with typical variation (+0.2 SD)- All functional networks demonstrate principal gradient values within expected ranges.

\subsection{Graph-Theoretic Descriptors}
Graph metrics characterize the topological structure of the brain’s functional network, quantifying how efficiently information flows and how modular or integrated the system is.

\subsubsection*{ROI-Level}
\begin{itemize}
    \item \textbf{Network Strength}: Weighted degree (sum of edge weights) for each of the 
    seven networks.  
    \textit{Measures hub-like activity and overall connectedness of each network.}
\end{itemize}

\subsubsection*{Global-Level}
\begin{itemize}
    \item \textbf{Modularity}: Degree of segregation into distinct functional modules.  
    \textit{Measures community structure and functional specialization.}

    \item \textbf{Global Efficiency}: Capacity for parallel information transfer between 
    distributed brain regions.  
    \textit{Indexes the overall integrative efficiency of the brain network.}

    \item \textbf{Average Clustering Coefficient}: Tendency for neighboring nodes to form 
    tightly interconnected clusters.  
    \textit{Captures local segregation and community tightness.}
\end{itemize}

\subsubsection*{Example:} The brain's functional architecture exhibits the following graph theoretical characteristics:- Modularity: Network segregation patterns fall within the expected range relative to the reference population.- Global Efficiency: Information integration capacity across the entire brain appears typical.- Clustering Coefficient: Local connectivity clustering remains within normative bounds.- Regional Connectivity: At the nodal level, the Ventral Attention network demonstrates increased functional coupling strength.

\subsection{ICA-Derived Descriptors}
ICA features capture the temporal dynamics and intrinsic activity patterns of independent large-scale functional networks. The ICA is estimated with a pre-defined template, followed by cross-subject decomposition. We use another set of functional networks (different from the FC) to provide another aspect of functional information. These networks are: visual, default-mode, cognitive control, sensorimotor, subcortical, auditory, and cerebellum.

\subsubsection*{ROI-Level (Network-Level)}
\begin{itemize}
    \item \textbf{Network Temporal Amplitude}: Mean absolute activation magnitude within 
    each ICA-defined network.  
    \textit{Represents the overall activity level of each network.}

    \item \textbf{Network Temporal Variability}: Standard deviation of temporal fluctuations 
    in each network.  
    \textit{Captures dynamic stability vs.\ variability of network activity.}

    \item \textbf{Network Spectral Ratio}: Ratio of slow (0.01--0.1\,Hz) to fast 
    (0.1--0.25\,Hz) oscillations.  
    \textit{Reflects dominant timescale of spontaneous network dynamics.}

    \item \textbf{Network Autocorrelation}: Lag-1 temporal coherence per network.  
    \textit{Measures persistence vs.\ rapid transitions in network activation.}

    \item \textbf{Network Transient Frequency}: Proportion of extreme activation events 
    ($|z| > 3$).  
    \textit{Indexes occurrence of transient bursts or network ``spikes''.}

    \item \textbf{Network-Pair FNC}: Functional network connectivity among ICA-derived 
    networks.  
    \textit{Represents coherence between independent components.}

    \item \textbf{Network fALFF}: Fractional amplitude of low-frequency fluctuations per 
    network.  
    \textit{Measures relative contribution of spontaneous low-frequency oscillatory activity.}
\end{itemize}

\subsubsection*{Global-Level}
\begin{itemize}
    \item \textbf{Overall Network Engagement}: Mean absolute activation across all ICA networks.  
    \textit{Summarizes global activity level.}

    \item \textbf{Overall Temporal Dynamics}: Average variability of network activity.  
    \textit{Indexes brain-wide dynamic stability vs.\ fluctuation.}

    \item \textbf{Overall Spectral Balance}: Global ratio of slow to fast oscillatory power.  
    \textit{Reflects dominant timescale of whole-brain spontaneous activity.}

    \item \textbf{Overall Temporal Coherence}: Average lag-1 autocorrelation across networks.  
    \textit{Represents persistence of whole-brain states.}

    \item \textbf{Overall Transient Activity}: Frequency of intense activation bursts across 
    the brain.  
    \textit{Measures prevalence of transient whole-brain activation events.}
\end{itemize}

\subsubsection*{Example:} The subject's independent component analysis reveals these functional network characteristics:- Temporal Dynamics: ICA timecourses demonstrate typical overall network engagement, standard temporal variability, prominent slow oscillation dominance, characteristic temporal coherence, and expected transient activity. At the network level, the Cerebellar system exhibits strong engagement in absolute mean amplitude and dynamic fluctuations in temporal variability. The Default Mode network displays slow oscillations in spectral ratio, while the Auditory network shows persistent activity in lag-1 autocorrelation. The Visual network demonstrates few transient events in outlier frequency.- Functional Coupling Patterns: Notable connectivity patterns include elevated coupling between Visual and Cognitive Control networks, within the Cerebellar network, and within the Default Mode network. Reduced connectivity is observed between Subcortical and Cognitive Control networks, and between Default Mode and Cerebellar networks.- Fractional ALFF: All networks show fractional amplitude of low-frequency fluctuations within typical ranges, suggesting balanced contributions across systems.

\subsection{High-Level Semantic Descriptors}

The high-level semantic descriptors cover the demographic, phenotypic, cognitive, and biomarker-related information of each subject. We listed the descriptors used in stage 3 in \cref{tab:semantic_desc} as well as what they measure and the available datasets.

\begin{table}[htbp]
\label{tab:semantic_desc}
\renewcommand{\arraystretch}{1.50}  
\setlength{\abovecaptionskip}{3pt}
\centering
\resizebox{\textwidth}{!}{
\begin{tabular}{c|c|c}
\hline
Name                                            & Meaning                                                                                                                                       & Avail Datasets   \\ \hline
Sex                                             & Male/Female                                                                                                                                   & All datasets     \\
Age                                             & age of subject in years                                                                                                                       & All datasets     \\
BMI                                             & Body Mass Index; an estimate of body fat based on weight relative to height.                                                                  & UKB, HCP, HCP-A, \\
Blood Pressure                                  & A measure of the force of blood against artery walls, given as diastolic values                                                               & UKB, HCP-A, ABCD \\
Cholesterol Level                               & Blood lipid profile indicating levels of total cholesterol; a marker of cardiovascular and metabolic health.                                  & UKB, HCP-A       \\
Fluid Intelligence Score                        & A measure of problem-solving ability, reasoning, and the capacity to think flexibly without relying on prior knowledge.                       & UKB              \\
Fluid Composite Score                           & A combined measure of fluid cognitive abilities in the Human Connectome Project, capturing reasoning, abstraction, and novel problem solving. & HCP, HCP-A, ABCD \\
Flanker Score                                   & An assessment of attention and inhibitory control; measures the ability to suppress distracting information.                                  & HCP-A            \\
APOE4 Status                                    & Genetic marker indicating presence of the APOE $\epsilon$4 allele, associated with increased Alzheimer’s disease risk.                                 & ADNI             \\
AV45 (Florbetapir PET SUVR)                     & A PET imaging biomarker of $\beta$-amyloid deposition; higher values indicate greater amyloid burden.                                               & ADNI             \\
CDRSB (Clinical Dementia Rating – Sum of Boxes) & A clinician-rated measure of cognitive and functional impairment severity across multiple domains.                                            & ADNI             \\
MMSE (Mini-Mental State Examination)            & A brief standardized test of global cognitive function, including memory, attention, and orientation.                                         & ADNI             \\
Verbal IQ                                       & A measure of verbal reasoning, vocabulary knowledge, and language-based cognitive abilities.                                                  & ADHD200, ABIDE2  \\
Performance IQ                                  & A measure of nonverbal reasoning, visual–spatial processing, and perceptual problem-solving skills.                                           & ADHD200, ABIDE2  \\ \hline
\end{tabular}
}
\end{table}

\FloatBarrier

\section{Details on Datasets}
\label{appendix:datasets}

The experimental targets for each dataset are detailed in Table \ref{tab:label}, where "Cls" denotes a classification task and "Reg" denotes a regression task. In line with the methods described in the previous section, we discretized two continuous variables. The fluid intelligence status was created by binning the z-scores of fluid intelligence relative to the UKB dataset. Similarly, the fluid composite status was formed by discretizing the corresponding z-scores relative to the HCP and HCP-A datasets.

\begin{table}[htbp]
\caption{Targets used in this study.}
\label{tab:label}
\renewcommand{\arraystretch}{1.50}  
\setlength{\abovecaptionskip}{3pt}
\centering
\resizebox{\textwidth}{!}{
\centering
\begin{tabular}{c|c|c|c}
\hline
Dataset      & Name                               & Task Type & Range/Possible Values                                                                                             \\ \hline
All Datasets & Sex                                & Cls       & Male/Female                                                                                                       \\
All Datasets & Age                                & Reg       & Float, 10-100                                                                                                     \\
All Datasets & Age Group                          & Cls       & adolescent ($x<18$), young adult ($18<x<30$), middle-aged adult ($30<x<60$), senior ($60<x<80$), elderly ($80<x$) \\
UKB          & Fluid Intelligence Score           & Reg       & Integer, 50-150                                                                                                   \\
UKB          & Fluid Intelligence Status          & Cls       & higher than usual ($z>1.5$),average ($-1.5<z<1.5$), lower than usual ($z<-1.5$)                                   \\
HCP,HCP-A    & Fluid Composite Score              & Reg       & Integer, 50-150                                                                                                   \\
HCP,HCP-A    & Fluid Composite Status             & Cls       & higher than usual ($z>1.5$),average ($-1.5<z<1.5$), lower than usual ($z<-1.5$)                                   \\
ADNI         & Alzheimer's Diagnosis              & Cls       & Cognitive Normal, Mild Cognition Impairment, Alzheimer\'s                                                         \\
ADNI         & APOE4 status                       & Cls       & APOE4 positive, APOE4 negative                                                                                    \\
ADHD200      & ADHD Diagnosis                     & Cls       & Control, ADHD                                                                                                     \\
ABIDE2       & Autism Spectrum Disorder Diagnosis & Cls       & Control, Autism                                                                                                   \\ \hline
\end{tabular}
}
\end{table}

\FloatBarrier

\section{Detailed Experimental Settings}

\begin{table}[htbp]
\caption{Hyperparameters of stage 1 training.}
\label{tab:hyper_stage1}
\centering
\resizebox{0.65\textwidth}{!}{
\centering
\begin{tabular}{ccc} \hline
\multicolumn{2}{c|}{Hyperparameters}                                                                  & Values       \\ \hline
\multirow{6}{*}{fMRI Tokenizer (small/base/large)} & \multicolumn{1}{c|}{Transformer encoder layers} & 12/12/24     \\
                                                   & \multicolumn{1}{c|}{Transformer decoder layers} & 12/12/24     \\
                                                   & \multicolumn{1}{c|}{Patch size}                 & 32/32/32     \\
                                                   & \multicolumn{1}{c|}{Embedding dimension}        & 384/768/1024 \\
                                                   & \multicolumn{1}{c|}{Num heads}                  & 6/12/16      \\
                                                   & \multicolumn{1}{c|}{Codebook size}              & 8192         \\ \hline
\multicolumn{2}{c|}{Batch size}                                                                      & 8            \\
\multicolumn{2}{c|}{Learning rate}                                                                   & 1e-4         \\
\multicolumn{2}{c|}{Minimal learning rate}                                                           & 1e-5         \\
\multicolumn{2}{c|}{Learning rate scheduler}                                                         & Cosine       \\
\multicolumn{2}{c|}{Optimizer}                                                                       & AdamW        \\
\multicolumn{2}{c|}{Total epochs}                                                                    & 50           \\ \hline
\end{tabular}
}
\end{table}

\begin{table}[htbp]
\caption{Hyperparameters of stage 2 training.}
\label{tab:hyper_stage2}
\centering
\resizebox{0.3\textwidth}{!}{
\centering
\begin{tabular}{c|c} \hline
Hyperparameters         & Value   \\ \hline
Batch size              & 16      \\ 
Learning rate           & 6e-4    \\
Minimal learning rate   & 6e-5    \\
Learning rate scheduler & Cosine  \\
Optimizer               & AdamW   \\
Total epochs            & 25      \\
Gradient clip           & 1.0     \\
Deepspeed stage         & stage 2 \\ \hline
\end{tabular}
}
\end{table}

\begin{table}[htbp]
\caption{Hyperparameters of stage 3 training.}
\label{tab:hyper_stage3}
\centering
\resizebox{0.35\textwidth}{!}{
\begin{tabular}{c|c}
\hline
Hyperparameters         & Value               \\ \hline
Batch size              & 32, 128 (zero-shot) \\
Learning rate           & 5e-4                \\
Minimal learning rate   & 5e-5                \\
Learning rate scheduler & Cosine              \\
Optimizer               & AdamW               \\
Total epochs            & 20                  \\
Gradient clip           & 1.0                 \\
Deepspeed stage         & stage 2             \\ \hline
\end{tabular}
}
\end{table}

\FloatBarrier

\section{Instruction Tuning Tasks and Paradigms}

\subsection{Single-question Single-answer Paradigm}

\begin{formal}
\#\#\# Question: \{QUESTION\}. \#\#\# Answer: \{ANSWER\}
\end{formal}

Here, \texttt{QUESTION} depends on the dataset and target variable. For example, for UK Biobank (UKB) sex prediction:
\begin{formal}
\#\#\# Question: What is the sex of this subject? \#\#\# Answer: Male
\end{formal}

To enhance robustness and reduce overfitting to specific phrasings, 200 distinct rewrites of each \texttt{QUESTION} type are generated to increase linguistic diversity.

\FloatBarrier

\subsection{Multi-question Multi-answer Paradigm}

For a single fMRI scan, multiple attributes may be queried simultaneously, each requiring an independent answer. The general template is:
\begin{formal}
\#\#\# Question: \{QUESTION1\}, \{QUESTION2\}. \#\#\# Answer: \{ANSWER1\} \{ANSWER2\}
\end{formal}

The number of questions and answers is flexible. For example, in UKB sex and fluid-intelligence prediction:
\begin{formal}
\#\#\# Question: What is the sex of this subject? How about its fluid intelligence status? \#\#\# Answer: Male \texttt{|} Below average
\end{formal}

Answers are separated using the ``\texttt{|}'' delimiter. Question rewrites also include fused forms in which multiple questions are asked within a single sentence:
\begin{formal}
\#\#\# Question: Can you specify the subject's sex and level of fluid intelligence (lower, mean, higher) from the fMRI scan? \#\#\# Answer: Male \,|\, Below average
\end{formal}

\FloatBarrier

\subsection{Open-Ended Question Paradigm}
In the open-ended setting, a broad question is posed without restricting the answer to any predefined set of targets. An example is:
\begin{formal}
\#\#\# Question: What specifics about the subject can you infer from the fMRI scan, such as demographics, cognitive abilities, or disease information? \#\#\# Answer: This is a senior male subject who is likely to have Alzheimer’s disease, with a positive APOE4 biomarker.
\end{formal}

To quantitatively assess correctness, a structured target schema is used, such as:
\begin{formal}
\{Sex: Male; Age Group: Senior; AD Diagnosis: AD; APOE4: Positive\}
\end{formal}
This structured reference is provided to an LLM-based evaluator, which scores each field for correctness. An overall correctness score is additionally computed, defined as correct only if all fields match the ground truth.

\FloatBarrier

\section{Results on Discretized Targets}
To better align with the language-modeling objective and enable more robust classification capabilities, several continuous or composite targets (e.g., age, fluid intelligence, and flanker performance) were discretized into categorical groups. Table~\ref{tab:discrete} reports the performance of fMRI-LM compared with two strong baselines (SWiFT and Brain-JEPA) across these discretized tasks.

\begin{table}[htbp]
\caption{Performance of fMRI-LM (base, GPT2) and baselined trained on discretized targets.}
\label{tab:discrete}
\centering
\resizebox{\textwidth}{!}{
\begin{tabular}{c|cc|cc|cc|cc|cc|cc}
\hline
           & \multicolumn{2}{c|}{UKB (age group)} & \multicolumn{2}{c|}{UKB (fluid intel status)} & \multicolumn{2}{c|}{HCP (fluid comp status)} & \multicolumn{2}{c|}{HCP-A (age group)} & \multicolumn{2}{c|}{HCP-A (fluid comp status)} & \multicolumn{2}{c}{HCP-A (flanker status)} \\ \cline{2-13} 
           & Acc               & F1               & Acc                   & F1                    & Acc                   & F1                   & Acc                & F1                & Acc                    & F1                    & Acc                  & F1                  \\ \hline
SWiFT      & 80.26             & 63.25            & 81.82                 & 30.24                 & 77.95                 & 37.24                & 55.28              & 39.05             & 96.21                  & 88.29                 & 83.22                & 61.05               \\
Brain-JEPA & 81.46             & 64.98            & 80.12                 & 35.29                 & 78.29                 & 30.25                & 70.19              & 48.74             & 93.28                  & 88.18                 & 88.91                & 62.19               \\
fMRI-LM    & 81.58             & 65.24            & 82.95                 & 37.54                 & 85.42                 & 38.12                & 69.52              & 48.29             & 97.10                  & 89.12                 & 89.65                & 64.26               \\ \hline
\end{tabular}
}
\end{table}

\FloatBarrier
\section{Results on Targets Independently Trained per Dataset}
Across the datasets used in this study, several targets are shared or aligned across sites. For example, UKB, HCP, and HCP-A all include sex classification; the UKB fluid-intelligence status corresponds closely to the fluid-composite status in HCP and HCP-A; and the diagnostic labels for control subjects overlap across ADNI, ADHD200, and ABIDE-II. Such overlap introduces potential sources of interference when these targets are trained jointly in a multi-task setting—particularly when datasets differ in sample size, demographic composition, or preprocessing pipelines.

To isolate the effect of multi-task training from dataset-specific signal and to better understand how fMRI-LM performs on each target, we train separate models independently on each dataset–target pair and compare them with the jointly trained multi-target model. This experiment allows us to assess (i) whether joint training introduces positive transfer, negative interference, or neither, and (ii) how consistently fMRI-LM generalizes when trained on focused versus shared objectives.

Table~\ref{tab:addition_perform} summarizes the performance differences between the jointly trained model and the independently trained models across all targets. Results are reported in terms of accuracy and AUC (when available). In the table, joint training generally maintains or slightly improves performance for UKB and HCP fluid-related tasks, suggesting positive transfer from shared cognitive phenotypes. However, for HCP-A, independent training yields higher accuracy in fluid-composite and flanker tasks, indicating dataset-specific specialization may be beneficial for older-adult cohorts. For clinical datasets, joint training provides modest improvements for ADNI and ABIDE-II, whereas ADHD200 shows minimal difference between settings. These results suggest that diagnostic targets benefit from shared representational features learned across diverse datasets, despite label heterogeneity.

Overall, the results demonstrate that multi-task joint training rarely harms performance, improves several clinical and cognitive targets, and preserves competitive performance on sex classification across datasets.

\begin{table}[htbp]
\caption{Performance of fMRI-LM (base, GPT2) trained independently on each target and jointly on all targets.}
\label{tab:addition_perform}
\centering
\resizebox{\textwidth}{!}{
\begin{tabular}{c|cc|cc|cc|cc|cc}
\hline
            & \multicolumn{2}{c|}{UKB (sex)}                 & \multicolumn{2}{c|}{UKB (fluid intel status)} & \multicolumn{2}{c|}{HCP (sex)} & \multicolumn{2}{c|}{HCP (fluid comp status)} & \multicolumn{2}{c}{HCP-A (sex)}  \\ \cline{2-11} 
            & Acc                    & AUC                   & Acc                   & AUC                   & Acc            & AUC           & Acc                   & AUC                  & Acc             & AUC            \\ \hline
Joint       & 94.89                  & 94.90                 & 82.95                 & -                     & 89.58          & 89.13         & 85.42                 & -                    & 89.58           & 88.98          \\
Independent & 94.72                  & 94.73                 & 83.03                 & -                     & 78.74          & 79.14         & 85.83                 & -                    & 86.81           & 87.01          \\ \hline
            & \multicolumn{2}{c|}{HCP-A (fluid comp status)} & \multicolumn{2}{c|}{HCP-A (flanker status)}   & \multicolumn{2}{c|}{ADNI (AD)} & \multicolumn{2}{c|}{ADHD200 (ADHD)}          & \multicolumn{2}{c}{ABIDE2 (ASD)} \\ \cline{2-11} 
            & Acc                    & AUC                   & Acc                   & AUC                   & Acc            & AUC           & Acc                   & AUC                  & Acc             & AUC            \\ \hline
Joint       & 97.10                  & 97.25                 & 89.65                 & 89.74                 & 77.92          & 79.91         & 72.92                 & 68.72                & 65.97           & 68.72          \\
Independent & 95.41                  & 96.02                 & 92.11                 & 93.05                 & 70.83          & 71.52         & 71.35                 & 70.15                & 59.90           & 70.21          \\ \hline
\end{tabular}
}
\end{table}

\FloatBarrier

\section{Further Results on Ablations}

\subsection{Ablation on weights of stage 1 and 2's loss terms}

\begin{figure}[htbp]
    \centering
    \label{fig:ablate_loss}
    \includegraphics[width=\linewidth]{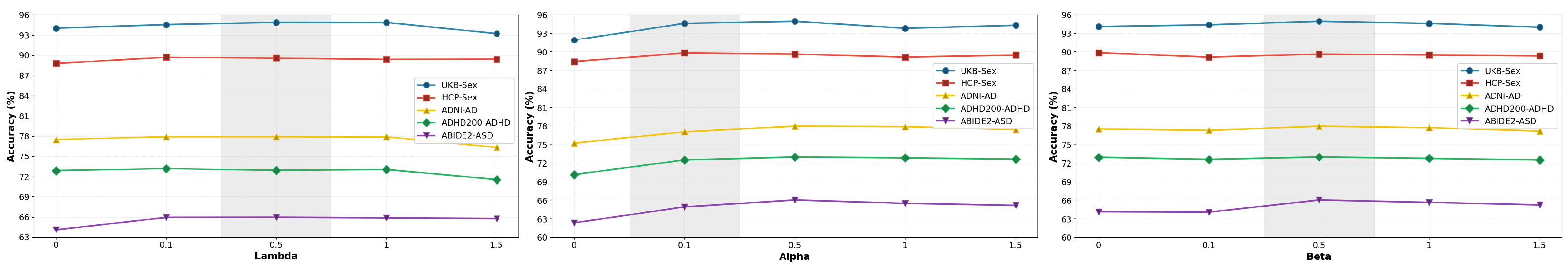}
    \caption{Ablations on the weights of loss terms in stage 1 and 2. The values in the main experiments are marked in gray. The  Changes in values of $\lambda, \alpha, \beta$ have no notable effect in the final performance, while changing them to 0 can have a negative effect.}
    \label{fig:placeholder}
\end{figure}

Here we present the performance of fMRI-LM on different loss weights, including $\lambda$ \cref{eq:total}  and $\alpha,\beta$ of \cref{eq:llm_total}. The results are in \cref{fig:ablate_loss}.

\FloatBarrier
\subsection{Ablation on different sizes of fMRI tokenizers and LLMs}
\begin{figure}[htbp]
    \centering
    \label{fig:ablate_size}
    \includegraphics[width=0.8\linewidth]{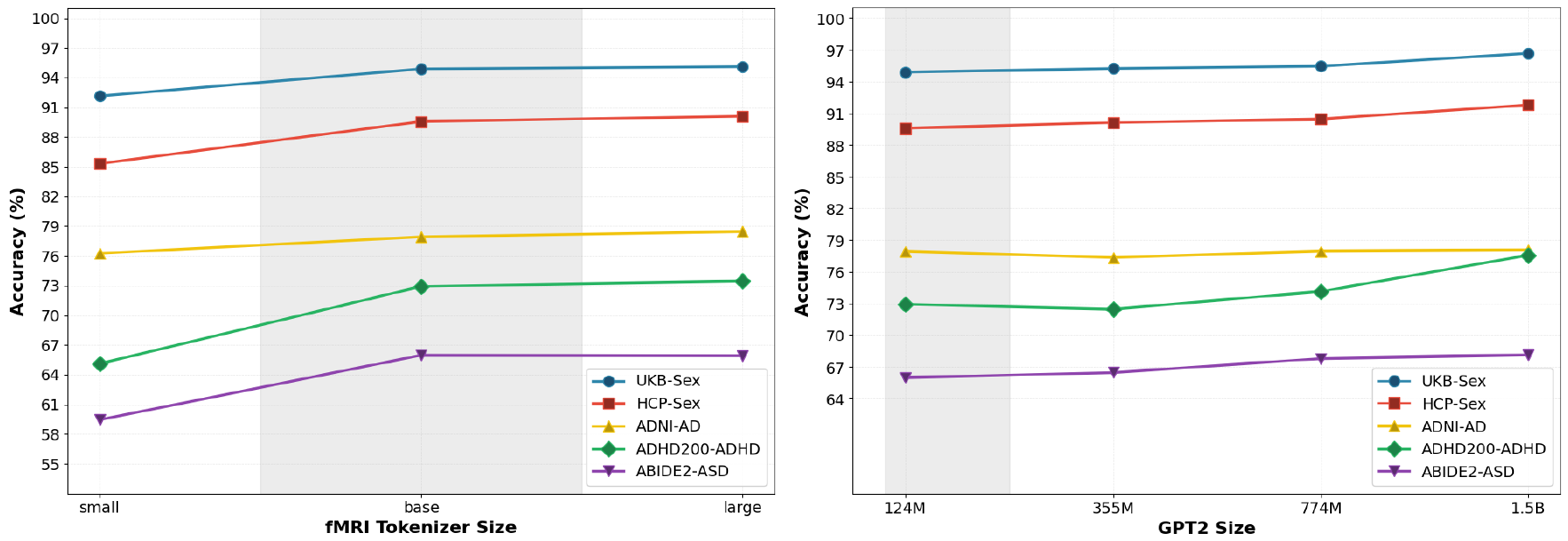}
    \caption{Compare performance with different sizes of fMRI tokenizers and GPT2 model.}
    \label{fig:placeholder}
\end{figure}

We further show the performances using different sizes of fMRI tokenizers (small, base, large) and LLMs (GPT2, from 124M to 1.5B). The results are in \cref{fig:ablate_size}.

\FloatBarrier

\section{Limitations and Future Work}
In this work, we align a universal fMRI representation with a large language model (LLM). To compensate for the lack of natural fMRI--text pairs, we construct large-scale fMRI descriptors derived from imaging-based features and fine-tune the LLM to interpret these representations for diverse downstream applications. A key limitation is that we do not yet fully capture or evaluate the model’s reasoning process over fMRI data. Ideally, the LLM would be able to explicitly identify the fMRI substrates (e.g., specific functional networks or regions) that support a given diagnosis or prediction, providing more transparent and mechanistic explanations. Achieving this will require more rigorous construction of instruction-tuning datasets, including carefully curated and validated annotations by domain experts, to minimize factual errors and support reliable, interpretable reasoning over brain signals. On the other hand, such imaging biomarkers are not fully revealed in the medical literature, so it will be difficult to examine the correctness of the reasoning.

\end{document}